\documentclass[runningheads]{llncs}
\usepackage{graphicx}
\usepackage{comment}
\usepackage{amsmath,amssymb} 
\usepackage{color}

\usepackage{times}
\usepackage{epsfig}
\usepackage{graphicx}
\usepackage{amsmath}
\usepackage{array}
\usepackage{hhline}
\usepackage{colortbl}
\usepackage{amssymb}
\usepackage{stfloats}
\usepackage{multirow}
\usepackage{booktabs}
\usepackage{threeparttable}
\usepackage{cite}
\usepackage{float}
\usepackage{caption}
\usepackage{eso-pic}
\usepackage[
font=scriptsize 
]{subfig}

\usepackage[pagebackref=true,breaklinks=true,letterpaper=true,colorlinks,bookmarks=false]{hyperref}

\def\etal{\textit{et al}.}
\def\ie{\textit{i.e.}}
\def\eg{\textit{e.g.}}

\definecolor{mygray}{gray}{.9}
\newcommand{\red}[1]{\textcolor{red}{#1}}
\newcommand{\green}[1]{\textcolor{green}{#1}}
\newcommand{\blue}[1]{\textcolor{blue}{#1}}
\begin{document}
\pagestyle{headings}
\mainmatter
\def\ECCVSubNumber{3089}  

\title{TENet: Triple Excitation Network for \\Video Salient Object Detection} 

\titlerunning{TENet: Triple Excitation Network for Video Salient Object Detection}
%
\author{Sucheng Ren\inst{1} \and
Chu Han\inst{2} \and
Xin Yang\inst{3} \and
Guoqiang Han\inst{1} \and
Shengfeng He\thanks{Corresponding author (hesfe@scut.edu.cn).}\inst{1}\orcidID{0000-0002-3802-4644}}
\authorrunning{S. Ren et al.}
%
\institute{School of Computer Science and Engineering, South China University of Technology \and
Guangdong Provincial People’s Hospital, Guangdong Academy of Medical Sciences \and
Department of Computer Science and Technology, Dalian University of Technology
}
\maketitle
\sloppy

\begin{abstract}
	In this paper, we propose a simple yet effective approach, named Triple Excitation Network, to reinforce the training of video salient object detection (VSOD) from three aspects, spatial, temporal, and online excitations. These excitation mechanisms are designed following the spirit of curriculum learning and aim to reduce learning ambiguities at the beginning of training by selectively exciting feature activations using ground truth. Then we gradually reduce the weight of ground truth excitations by a curriculum rate and replace it by a curriculum complementary map for better and faster convergence. In particular, the spatial excitation strengthens feature activations for clear object boundaries, while the temporal excitation imposes motions to emphasize spatio-temporal salient regions. Spatial and temporal excitations can combat the saliency shifting problem and conflict between spatial and temporal features of VSOD. Furthermore, our semi-curriculum learning design enables the first online refinement strategy for VSOD, which allows exciting and boosting saliency responses during testing without re-training. The proposed triple excitations can easily plug in different VSOD methods. Extensive experiments show the effectiveness of all three excitation methods and the proposed method outperforms state-of-the-art image and video salient object detection methods.
\end{abstract}

\section{Introduction}
When humans look into an image or a video, our visual system will unconsciously focus on the most salient region. The importance of visual saliency has been demonstrated in a bunch of applications, \eg, image manipulation~\cite{manipulation1, manipulation2}, object tracking~\cite{object_tracking1}, person re-identification~\cite{people_reid1, people_reid2,people_reid3}, and video understanding~\cite{video_understanding1, video_understanding2, video_understanding3}. According to the slightly different goals, saliency detection can be further separated into two research interests, eye-fixation detection~\cite{eyefixation1, eyefixation2 } which mimics the attention mechanism of the human visual system, and salient object detection (SOD)~\cite{BASNET,SOD1,SOD2} which focuses on segmenting the salient objects. In this paper, we focus on the latter.

\begin{figure}[t]
    \centering
    \setlength{\tabcolsep}{.5pt}
    \renewcommand{\arraystretch}{.5}
    \begin{tabular}{cccccc}
        \includegraphics[scale=0.117]{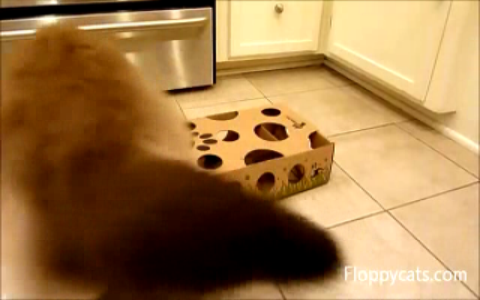}&
        \includegraphics[scale=0.117]{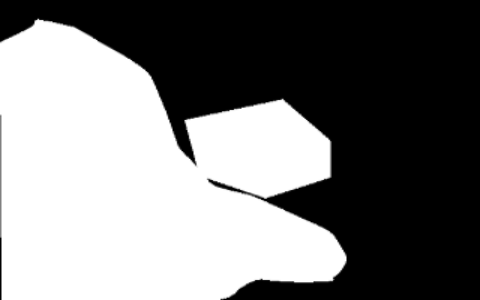}&
        \includegraphics[scale=0.117]{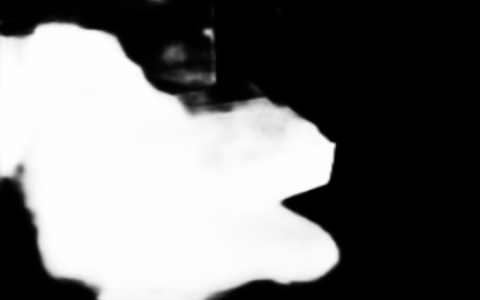}&
        \includegraphics[scale=0.117]{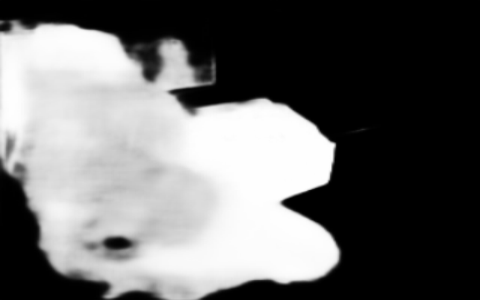}&
        \includegraphics[scale=0.117]{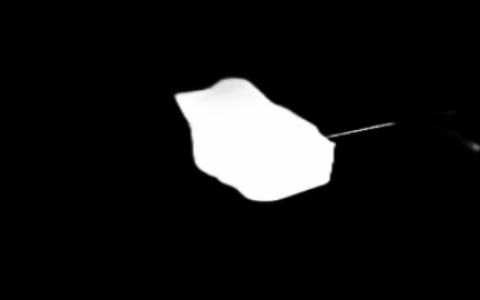}&
        \includegraphics[scale=0.117]{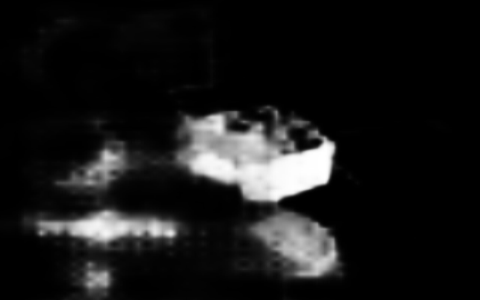}\\

        \includegraphics[scale=0.117]{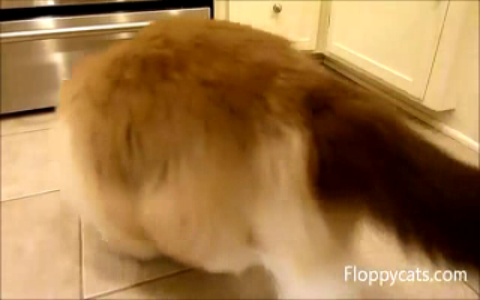}&
        \includegraphics[scale=0.117]{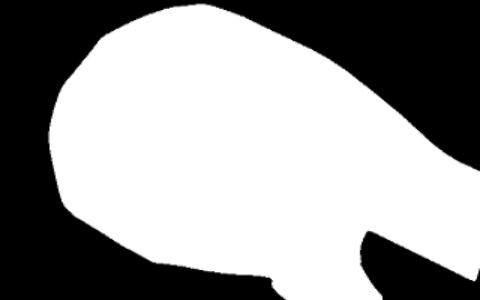}&
        \includegraphics[scale=0.117]{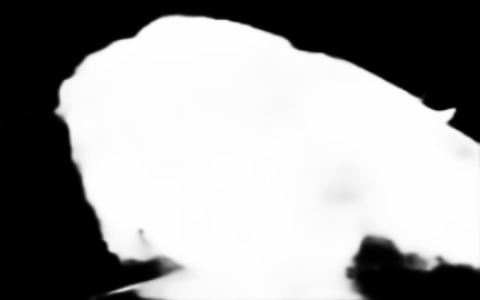}&
        \includegraphics[scale=0.117]{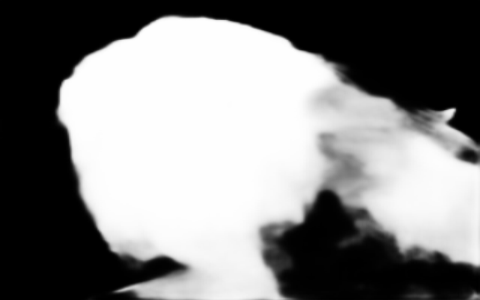}&
        \includegraphics[scale=0.117]{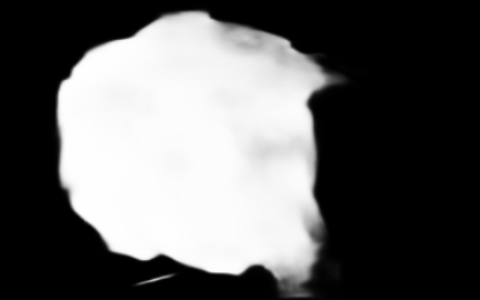}&
        \includegraphics[scale=0.117]{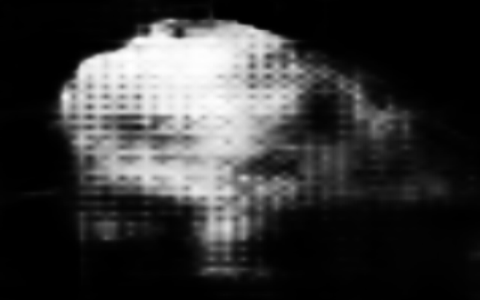}\\

        \includegraphics[scale=0.117]{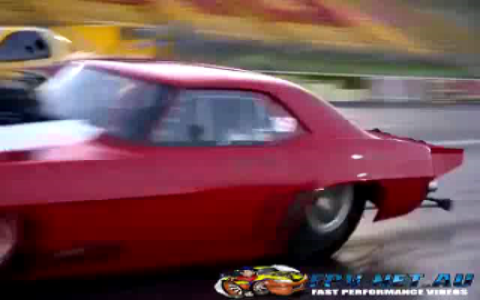}&
        \includegraphics[scale=0.117]{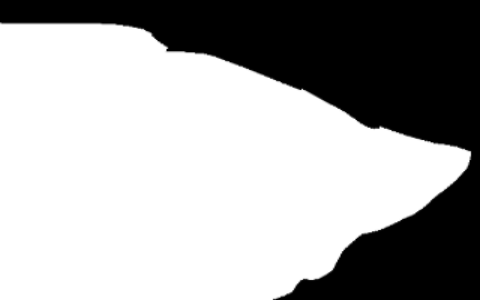}&
        \includegraphics[scale=0.117]{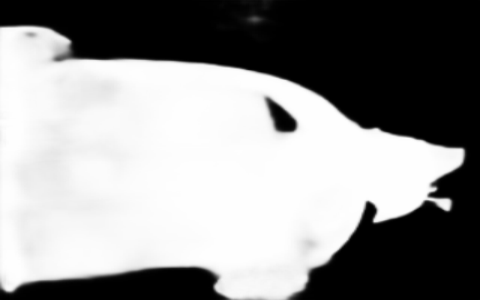}&
        \includegraphics[scale=0.117]{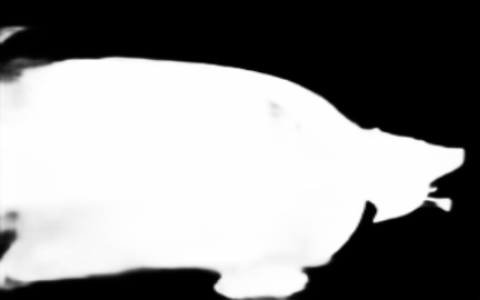}&
        \includegraphics[scale=0.117]{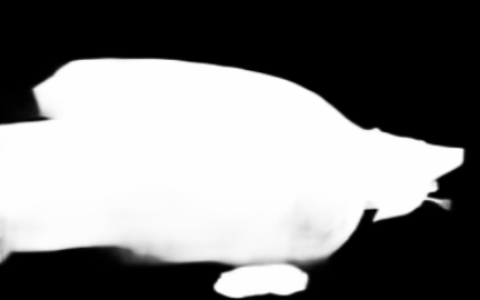}&
        \includegraphics[scale=0.117]{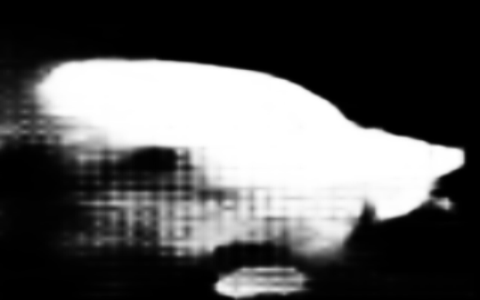}\\

        \includegraphics[scale=0.117]{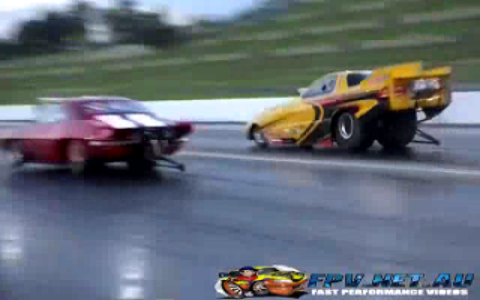}&
        \includegraphics[scale=0.117]{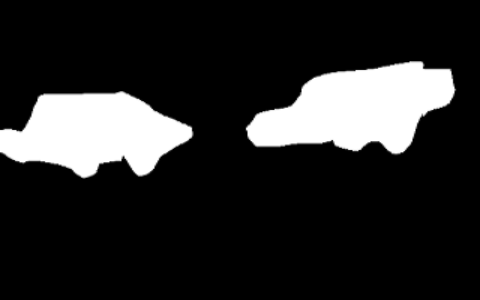}&
        \includegraphics[scale=0.117]{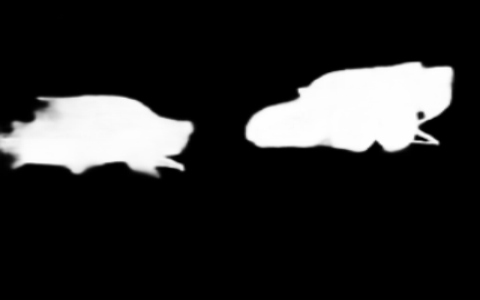}&
        \includegraphics[scale=0.117]{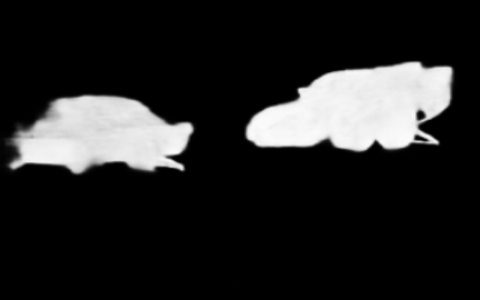}&
        \includegraphics[scale=0.117]{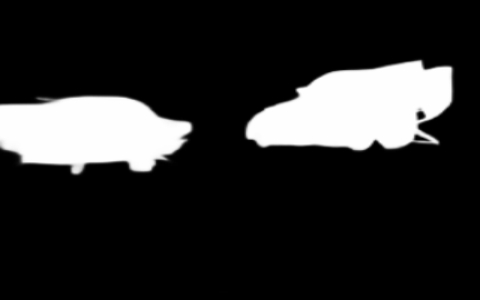}&
        \includegraphics[scale=0.117]{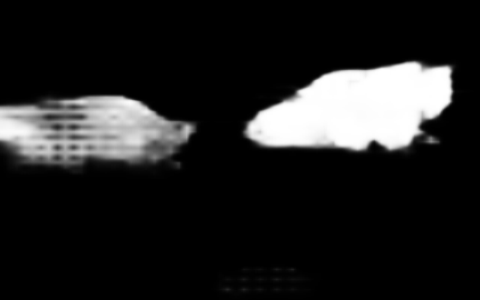}\\

        Input & GT & Ours & Ours w/o & BASNet & SSAV \\
        & &  & excitation & &
    \end{tabular}\vspace{-3mm}
    \caption{We propose to manually excite feature activations from three aspects, spatial and temporal excitations during training, and online excitation during testing for video salient object detection. Our simple yet effective solution injects additional spatio-temporal guidance during training and even testing for better and faster convergence. In contrast, the image-based method BASNet~\cite{BASNET} lacks temporal understanding (first row), while the video-based method SSAV~\cite{Fan_2019_CVPR} suffers from spatially coarse results.}
    \label{img:First_Image}\vspace{-5mm}
\end{figure}

Image-based salient object detection~\cite{BASNET, SOD2} has been made great achievements recently. However, detecting salient objects in videos is a different story. This is because the human visual system is influenced not only by appearance but also by motion stimulus. Therefore the severe saliency shifting problem~\cite{attention_shift1, attention_shift2, attention_shift3} poses the challenge in video salient object detection (VSOD). Despite different cues, \eg, optical flow~\cite{Flow_encoder, Pyramid, Fan_2019_CVPR} and eye-fixation~\cite{Fan_2019_CVPR}, are used to deal with this problem, the sudden shift of ground truth label makes the training difficult to converge.

Another issue in VSOD training is the contradictory features in spatial and temporal domains. As motion stimulus is a key factor of the human visual system, humans may pay attention to a moving object that does not distinct in appearance. Although features fusion is typically applied, extracting temporal features is much more difficult than spatial ones, as motion blurring, object, and camera movements are involved. They cannot capture clear object boundaries as the spatial features do. As a consequence, VSOD methods (\eg, the last column of Fig.~\ref{img:First_Image}) produce spatially coarse results in the scenarios with moving objects. We argue that a simple feature fusion strategy cannot solve this problem, and alternative guidance during training is desired.

To address the above two issues, we propose a Triple Excitation Network (TENet) for video salient object detection. Three types of excitations are tailored for VSOD, \ie, spatial and temporal excitations during training, and online excitation during testing. We adopt a similar spirit with curriculum learning \cite{Bengio:2009}, that we aim to loosen the training difficulties at the beginning by exciting selective feature activations using ground truth, then gradually increase task difficulty by replacing such ground truth by our learnable complementary maps. This strategy simplifies the training process of VSOD and boosts the performance with faster convergence and we name it as semi-curriculum learning. In particular, spatial excitation learns spatial features to obtain a boundary-sharp segment. While the temporal excitation aims to leverage previous predictions and excites spatio-temporal salient regions from a spatial excitation map and an optical flow. These excitations are directly performed on the activations of features, and therefore provide direct supports on mitigating errors brought by the problems of saliency shifting and inaccurate temporal features. Thanks to our semi-curriculum learning design, we can apply excitations in testing by proposing online excitation which can be done without any further training. It is worth noting that the proposed excitation mechanism can easily plug in different VSOD methods. Extensive experiments are performed to qualitatively and quantitatively evaluate the effectiveness of the proposed method, it outperforms state-of-the-art methods on four VSOD benchmarks.

The main contributions of this paper are four-fold:
\begin{itemize}
    \item We delve into the problems of saliency shifting and inaccurate temporal features and tailor a triple excitation mechanism to excite spatio-temporal features for mitigating the training difficulties caused by these two problems. Better and faster convergence can be obtained.

    \item We present a semi-curriculum learning strategy for VSOD. It reduces the learning ambiguities by exciting certain activations during the beginning of training, then gradually reduces the curriculum rate to zero and transfers the weight of excitation from ground truth to our learnable complementary maps. This learning strategy progressively weans the network off dependence on ground truth, which is not only beneficial for training but also for testing.

    \item We propose an online excitation that allows the network to keep self-refining during the testing phase.

    \item We outperform state-of-the-art SOD and VSOD methods on four benchmarks.
\end{itemize}


\section{Related Works}
\textbf{Image Salient Object Detection.}
Traditional image saliency detection methods~\cite{SIVM, MSTM, SFLR} usually rely on the hand-crafted features, \eg, color contrast, brightness. It can be separated into two categories, bottom-up~\cite{bottom1, bottom2, top-bottom} and top-down~\cite{top, top-bottom} approaches. Driven by a large amount of labeled data, researchers attempt to consider saliency detection as a classification problem~\cite{saliency_classification1, saliency_classification2} by simply classifying the patches into non-salient or salient. However, the patches cropped from the original image are usually small and lack of global information. Recent approaches adopt FCN~\cite{long2015fully} as a basic architecture to detect saliency in an end-to-end manner. Based on that, edge information is incorporated to promote clear object boundaries, by a boundary-enhanced loss~\cite{boundaryloss} or jointly trained with edge detection~\cite{SOD2}. Attention mechanism~\cite{Attention1, Attention2} is also introduced to filter out a cluttered background. All these methods provide a useful guideline to handle spatial information.


\textbf{Video Salient Object Detection.}
The involved temporal information makes video salient object detection much harder than image salient object detection. Some existing approaches try to fuse spatial and temporal information using graph cut~\cite{graph_cut1, graph_cut2}, gradient flow~\cite{ViSal}, and low-rank coherence~\cite{SFLR}. Researchers also try to associate spatial with temporal information using deep networks. Wang~\etal~\cite{VSOD_FCN} concatenate the current frame and saliency of the previous frame to process temporal information. Li~\etal~\cite{Flow_encoder} propose to use optical flow to guide the recurrent neural encoder. They use ConvLSTM to process optical flow and warp latent features before feeding into another ConvLSTM. To capture a wider range of temporal information, a deeper bi-directional ConvLSTM\cite{Pyramid} has been proposed. Fan~\etal~\cite{Fan_2019_CVPR} mitigate the saliency shifting problem by introducing the eye-fixation mechanism. Similar to VSOD, unsupervised video object segmentation~\cite{Wang_2019_CVPR,yang2019anchor,Lu_2019_CVPR} aims at segmenting primarily objects with temporal information. However, as discussed above, replying only to additional features cannot solve problems of saliency shifting and inaccurate temporal features well. We resolve them from the perspective of reducing training difficulties. 

\textbf{Extra Guidance for CNNs.}
Introducing extra guidance is a popular solution to aid the training of a deep network. For example, jointly training semantic segmentation and object detection~\cite{seg_detect1, seg_detect2, seg_detect3} improves the performances for both tasks. One limitation of multi-task training is that it needs two types of annotations. Some other works introduce two different types of annotations from the same task, \eg, box and segmentation labels for semantic segmentation~\cite{detect1, detect2}, to boost the training performance. Different from these methods, we do not introduce extra task or annotation for training, but directly employ ground truth of the same task as well as pseudo-label for exciting features activations.

\section{Triple Excitation Network}
\label{TENet}
\begin{figure*}[t]
\center
\includegraphics[width=\linewidth]{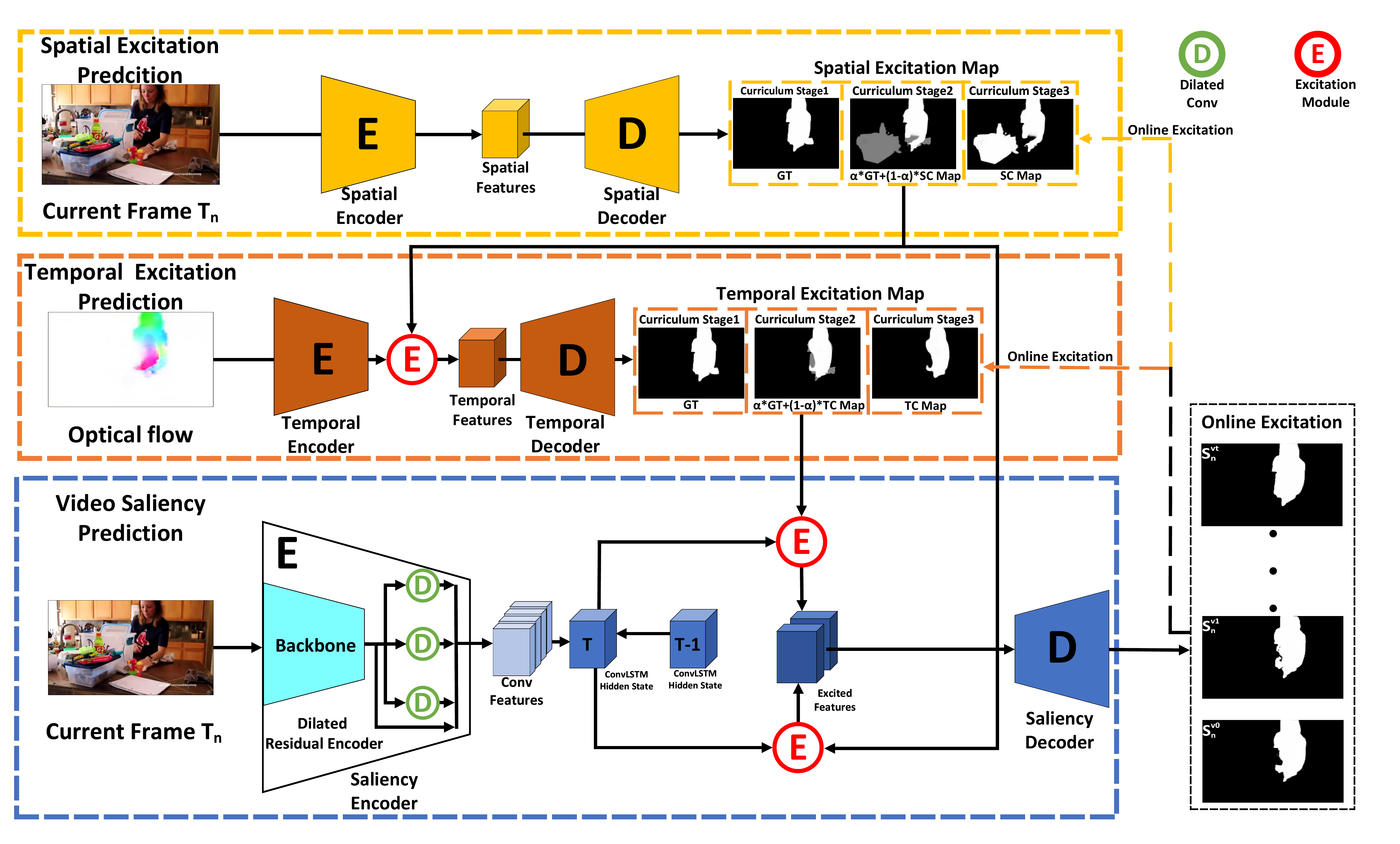}
\caption{Network architecture of TENet. The upper two branches provide spatial and temporal excitation with a curriculum learning strategy. In each curriculum stage, we balance the contribution of the ground truth and the complementary map to avoid the overdependency of ground truth during training phase. ConvLSTM is applied in the third branch to introduce the temporal features from the previous frames. During testing phase, an optional online excitation allows the network to keep refining the saliency prediction results by keeping updating the complementary maps with previous predictions recurrently. Note that, the structures of all the encoders (E) in the network are the exactly the same but with different parameters. We only show the saliency encoder for simplification.}
\label{fig:method}
\end{figure*}

\subsection{System Overview}
\label{sec:overview}
Given a series of frames $\{T_{n}| n = 1,2,..., N\}$, we aim to predict the salient object in frame $T_n$. Fig.~\ref{fig:method} shows the pipeline of our proposed Triple Excitation Network. The basic network is an encoder-decoder architecture with skip connections (which are hidden in Fig.~\ref{fig:method} for simplification). Our framework consists of three branches with respective purposes. The spatial excitation prediction branch is proposed to predict spatial complementary maps with rich spatial information for generating spatial excitation map. The temporal excitation prediction branch leverages the optical flow and spatial excitation maps to generate the temporal complementary maps. These two complementary maps are combined with ground truth to provide the additional guidance for the network training and testing. ConvLSTM~\cite{NIPS2015_5955} is introduced to inject temporal information on the feature maps extracted from the encoder in video saliency prediction branch. After the spatial and temporal excitations, the final saliency map of the current frame $T_n$ is generated by the saliency decoder. During the testing phase, the final saliency map is further adopted for online excitation.

\subsection{Excitation Module}
\label{sec:Excitation}
\begin{figure}[t]
	\center
	\includegraphics[width=.8\linewidth]{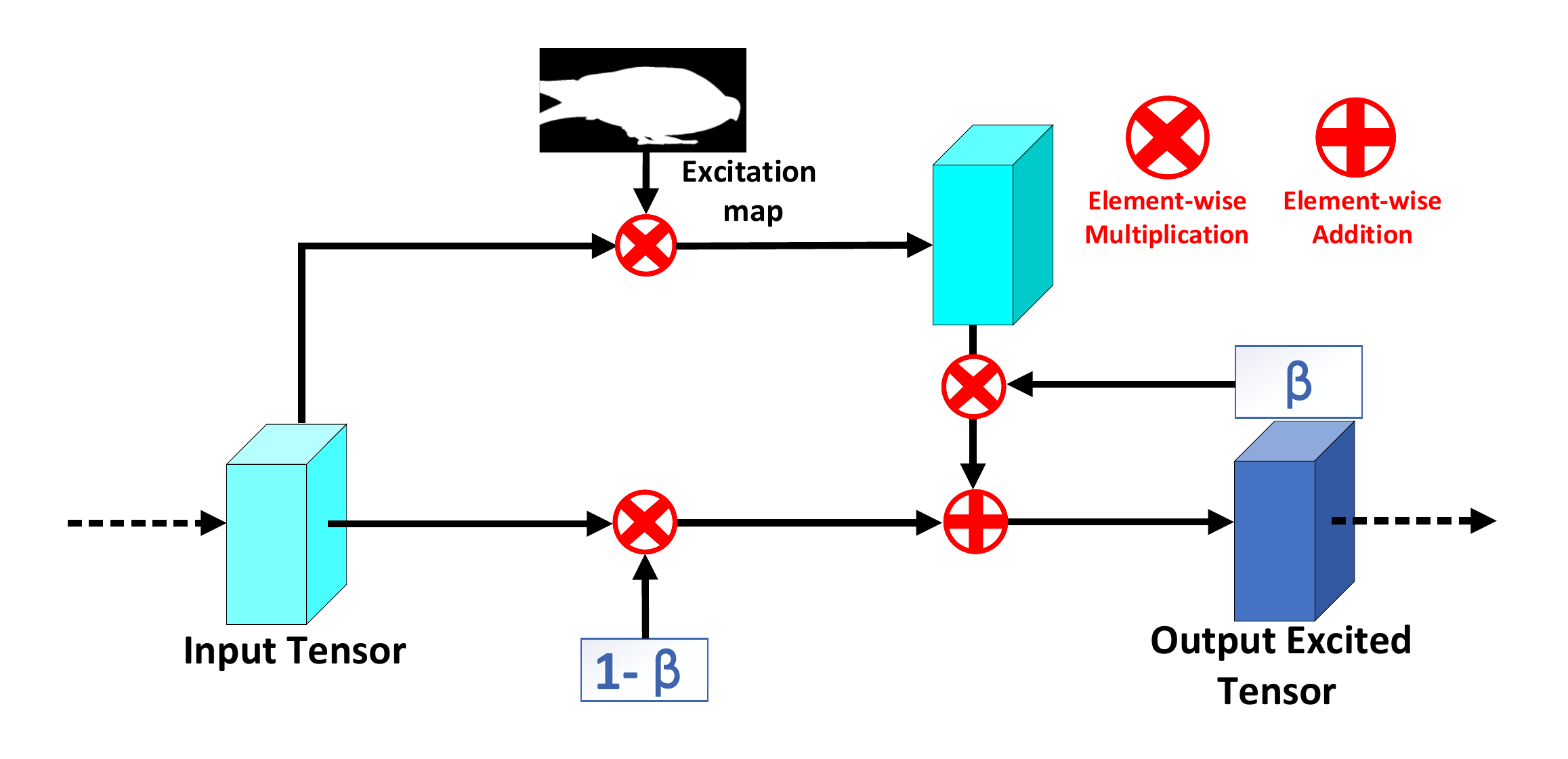}
	\caption{The proposed excitation module. It strengthens saliency features responses by manually exciting certain feature activations, and the training is controlled by a learnable excitation rate $\beta$ automatically adjusted according to the feedback of neural network.}
	\label{fig:excitation}
\end{figure}

Due to the difficulties of handling saliency shifting and the contradictory features in spatial and temporal domains, a simple feature fusion strategy is no longer sufficient for video salient object detection. Therefore, we propose an excitation mechanism shown in Fig.~\ref{fig:excitation} as the additional guidance to reinforce certain feature activations during training.

It is worth noting that our proposed excitation mechanism is a super lightweight plug-in that does not waste computational power because it does not involve any convolution operation. Given an input tensor $M$, an excitation map $E$ with the pixel values under the range $[0,1]$, we can obtain the output excited tensor $M'$ by the following equation:
\begin{equation}
    M' = \beta \odot E \odot M + (1-\beta) \odot M,
\end{equation}
where $\odot$ is element-wise multiplication. $\beta$ is a learnable excitation rate which determines the intensity of excitation based on the feedback of the model itself. It learns an optimum balance between manual excitation and learned activations.

\subsubsection{Semi-curriculum Learning.}
The excitation map can actually be any map that reflects the feature responses required excitation. It can be the ground truth saliency map in this application. However, directly utilizing the ground truth for excitation may let the network excessively rely on the ground truth itself. Therefore, we introduce a semi-curriculum learning strategy for our excitation mechanism. This strategy shares a similar spirit to the curriculum learning framework \cite{Bengio:2009}, in which they argue that training with an easy task at first then continues with more difficult tasks gradually may leads to better optimization. Therefore, as the training goes on, we update the excitation map by trading off the intensity between the ground truth $GT$ and a learnable complementary map $S$ as follows:
\begin{equation}
E = \alpha \odot GT + (1-\alpha) \odot \mathcal{S}
\label{eq:excitation_map}
\end{equation}
where $\alpha$ is the curriculum rate which has been initially set as 1 and will automatically decay to 0 to transfer the contribution from the ground truth to the learnable complementary map. The gradually decreased curriculum rate $\alpha$ will increase the task difficulty and thus can effectively avoid the overdependence on the ground truth. In the meanwhile, this is also the key to enable online excitation.

In practice, we divide the training process into three curriculum stages along with the change of training epoch $e$. The excitation map $E$ in Eq.~\ref{eq:excitation_map} can be reformulated as follows:
\begin{equation}
\label{eq:excitation_map2}
E =
\begin{cases}
GT & {\rm Stage 1:} e \leq 2\\
\alpha \odot GT + (1-\alpha) \odot \mathcal{S}& {\rm Stage 2:} 2 \textless   e \leq 10\\
\mathcal{S} & {\rm Stage 3:} e \textgreater 10
\end{cases}
\end{equation}

\textbf{Stage 1:} Due to the imbalance foreground and background pixels in VSOD, for example, the salient pixels take only 8.089\% in the DAVIS Dataset~\cite{DAVIS}, reinforcing the network to focus on salient region by using the ground truth as the excitation map provides a shortcut for a better optimization at the beginning of training.

\textbf{Stage 2:} However, models tend to rely on the perfect ground truth and it degrades the model performance once the ground truth is removed. As a result, we gradually replace the ground truth by our learned complementary map (controlled by the curriculum rate $\alpha$). During this period, the predicted complementary map is to inject perturbation and prevent the model from too sensitive to the perfect ground truth.

\textbf{Stage 3:} When $\alpha$ decays to zero, our model is excited only by the complementary maps. This avoids our network overdependence on GT, more importantly, this is the key to enabling online excitation. We keep training the network in this stage to 15 epochs.

\subsection{Spatial-temporal Excited Saliency Prediction}
\label{sec:sptail-temporal excitation}
Our model consists of three branches, the first two are for generating excitation maps, and we predict the video saliency result in the third branch by predicting the video frames one by one. For all these branches, we extract the feature maps by a dilated residual encoder as described below.

\subsubsection{Dilated Residual Encoder.}
\label{sec:encoder}
The backbone of the encoder borrows from ResNet~\cite{ResNet}. We replace the first convolutional layer and the following pooling layer by a 3$\times$3 convolutional layer with stride 1 and extract the deep features $\mathcal{X}_n \in \mathbb{R}^{w \times h \times c}$. To handle the uncertain scales of the objects, we extract the multi-level feature maps by introducing dilated convolution and keep increasing the dilation rates $\{ 2^k \}_{k=1}^K$. The $k$-th level features extracted from the dilated convolution with dilation rate $2^k$ is $\mathcal{D}^k \in \mathbb{R}^{w \times h \times c'}$. The output feature maps $\mathcal{F}$ is the concatenation of all the outputs from dilated residual encoder:
\begin{equation}
\mathcal{F}_n = \left[ \mathcal{X}_n, \mathcal{D}_n^1, \mathcal{D}_n^2,...,\mathcal{D}_n^K \right],
\end{equation}
where $\mathcal{F}$ $\in \mathbb{R}^{w \times h \times (c+K*c')}$. The feature maps $\mathcal{F}_n$ from dilated residual encoder not only keeps the original features $\mathcal{X}_n$ but also covers much larger receptive fields with local-global information. All the encoders in the three branches share the same structure but have different parameters.

\subsubsection{Spatial Excitation Prediction Branch.}
\label{sec:Excitation Map Prediction}
We predict the spatial excitation map from a single frame in the spatial excitation branch which has an encoder-decoder structure. We use the dilated residual encoder as mentioned above, and the decoder has four convolutional stages. Each stage contains three convolutional blocks, each of which is a combination of a convolutional layer, a batch normalization layer, and a ReLU activation layer. With the spatial complementary map $S_n^s$ generated by the spatial decoder, we can calculate the spatial excitation map $E_n^s$ by Eq.~\ref{eq:excitation_map2}.

\subsubsection{Temporal Excitation Prediction Branch.}
The temporal branch is designed to tackle the human visual attention shifting problem. It takes an input optical flow, which is calculated by the state-of-the-art optical flow prediction method~\cite{liu2019selflow}, and outputs a temporal excitation map. This branch has the same network structure with the spatial one. The difference is, we make a spatial excitation on the latent features in the temporal branch in order to associate the temporal excitation with the spatial excitation.

Given the input optical flow from the frame $T_{n-1}$ to the frame $T_n$, we have the latent features $\mathcal{F}^t_n$ extracted from the temporal encoder. Then the spatial excitation map $E^s_n$ is for spatial excitation. The excited temporal latent features $\mathcal{F'}_n^t$ is calculated as follows:
\begin{equation}
    \mathcal{F'}_n^t = \beta^{s \rightarrow t} \odot E^s_n \odot \mathcal{F}_n^t + (1-\beta^{s \rightarrow t}) \odot \mathcal{F}_n^t,
\end{equation}
where $\beta^{s \rightarrow t} $ is a learnable temporal excitation rate. The optical flow reveals the moving objects explicitly and the predicted temporal complementary map covers temporally salient regions for governing training.
With the temporal complementary map $\mathcal{S}_n^t$ generated by the temporal decoder, we can calculate the temporal excitation map $E_n^t$ by Eq.~\ref{eq:excitation_map2}.

\subsubsection{Video Saliency Prediction Branch.}
\label{sec:spatial}
In this branch, we aim to predict saliency maps with spatially sharpen the object boundaries by leveraging the spatio-temporal excitation mechanism. After the feature extraction with dilated residual encoder, We apply the bi-directional ConvLSTM on the extracted feature maps $\mathcal{F}^v_n$ to obtain long-short term spatial and temporal features:
\begin{equation}
    \overrightarrow{\mathcal{H}}_n^v = ConvLSTM(\mathcal{F}^v_n, \overrightarrow{\mathcal{H}}_{n-1}^v),
\end{equation}
\begin{equation}
\overleftarrow{\mathcal{H}}_n^v = ConvLSTM(\mathcal{F}^v_n, \overleftarrow{\mathcal{H}}_{n-1}^v).
\end{equation}
We consolidate the features representations on two temporal directions by leveraging both spatial excitation and temporal excitation and obtain bi-directional feature maps $\overrightarrow{\mathcal{H'}}^v_n$ and $\overleftarrow{\mathcal{H'}}^v_n$ of frame $T_n$ as follows:
\begin{equation}
\begin{split}
\overrightarrow{\mathcal{H'}}^v_n = cat(&\overrightarrow{\beta}^{s \rightarrow v} \odot E^s_n \odot \overrightarrow{\mathcal{H}}^v_n + (1-\overrightarrow{\beta}^{s \rightarrow v}) \odot \overrightarrow{\mathcal{H}}_{n}^v \\
+&\overrightarrow{\beta}^{t \rightarrow v} \odot E^t_n \odot \overrightarrow{\mathcal{H}}_{n}^v + (1-\overrightarrow{\beta}^{t \rightarrow v}) \odot \overrightarrow{\mathcal{H}}_{n}^v),
\end{split}
\end{equation}
\begin{equation}
\begin{split}
	\overleftarrow{\mathcal{H'}}^v_n = cat(&\overleftarrow{\beta}^{{s \rightarrow v}} \odot E^s_n \odot \overleftarrow{\mathcal{H}}^v_n + (1-\overleftarrow{\beta}^{s \rightarrow v}) \odot \overleftarrow{\mathcal{H}}^v_n \\
	+&\overleftarrow{\beta}^{t \rightarrow v} \odot E^t_n \odot \overleftarrow{\mathcal{H}}_n^v + (1-\overleftarrow{\beta}^{t \rightarrow v}) \odot \overleftarrow{\mathcal{H}}_n^v),
\end{split}
\end{equation}
where $E^s_n$ and $E^t_n$ are the spatial and temporal excitation maps respectively and $cat(\cdot, \cdot)$ is the concatenation operation. $\beta^\cdot$ are the learnable parameters. Since we perform the excitation strategy on the latent space, the excitation maps will be first downsampled to the same size with the feature maps.

The bi-directional excited hidden stage is then concatenated together to produce the final saliency result $\mathcal{S}^v_n$ of the frame $T_n$ by the saliency decoder $W^s$:
\begin{equation}
    \mathcal{S}^v_n = W^s \otimes cat(\overrightarrow{\mathcal{H'}}^v_n, \overleftarrow{\mathcal{H'}}^v_n).
\end{equation}
Note that the saliency decoder $W^s$ here has the same structure but different parameters with temporal and spatial decoders.

\subsection{Loss Function}
\label{sec:loss}
We borrow the loss function from BASNet~\cite{BASNET}. It includes the cross entropy loss~\cite{BCE}, SSIM loss~\cite{SSIM}, and IoU loss~\cite{iou}. They measure the quality of saliency map in pixel-level, patch-level, and object-level respectively.
\begin{equation}
l = l_{bce} + l_{ssim} + l_{IoU}.
\end{equation}

The cross entropy loss ${l}_{bce}$ measures the distance between two probability distributions which is the most common loss function in binary classification and salient object detection.
\begin{equation}
    l_{bce}(\mathcal{S}, GT) = -\sum\limits_{i=1}^w\sum\limits_{j=1}^h[GT(i, j) \log \mathcal{S}(i, j) + (1-GT(i, j))\log(1-\mathcal{S}(i,j))],
\end{equation}
where $S$ is the network predicted saliency map, $S_{gt}$ is the ground truth saliency map.

The SSIM is originally designed for measuring the structural similarity of two images. When applying this loss into saliency detection, it helps the network pay more attention to the object boundary due to the higher SSIM activation around the boundary. Let $x = \left\{ x_i:i=1,\cdots,N^2 \right\}, y = \left\{ y_i:i=1,\cdots,N^2 \right\}$ be the corresponding N$\times$N patches of predict saliency and ground truth label respectively, we have:
\begin{equation}
    l_{ssim}(\mathcal{S}, GT) = 1-\frac{(2\mu_x\mu_y+c_1)(2\sigma_{xy}+c_2)}{(\mu_x^2+\mu_y^2+c_1)(\sigma_x^2+\sigma_y^2+c_2)},
\end{equation}
where $\mu_x, \mu_y$ and $\sigma_x, \sigma_y$ are the mean and variance, and $\sigma_{xy}$ is the co-variance. $c_1=0.01^2,~c_2=0.03^2$ are constants for maintaining stability.

Intersection over Union (IoU) is widely used in detection and segmentation for evaluation and also used as training loss. The IoU loss is defined as:
\begin{equation}
    l_{iou}(\mathcal{S}, GT) = 1-\frac{\sum\limits_{i=1}^{w}\sum\limits_{j=1}^{h}\mathcal{S}(i,j)GT(i,j)}{\sum\limits_{i=1}^{w}\sum\limits_{j=1}^{h}[\mathcal{S}(i,j)+GT(i,j)-\mathcal{S}(i,j)GT(i,j)]}.
\end{equation}

The above losses apply to three branches, and the total objective function of our network is the combination of the spatial excitation loss $l(\mathcal{S}^s, GT)$, temporal excitation loss $l(\mathcal{S}^t, GT)$, and the video saliency loss $l(\mathcal{S}^v, GT)$:
\begin{equation}
\mathcal{L} = l(\mathcal{S}^s, GT) + l(\mathcal{S}^t, GT) + l(\mathcal{S}^v, GT).
\end{equation}

\subsection{Online Excitation}
\label{sec:Online}
In our network design, the quality of the excitation map plays an important role in final saliency map prediction. During training, we use a predicted excitation map for highlighting salient activations in the features. Thanks to our semi-curriculum learning strategy, the excitation map does not rely on GT during the testing phase, and we can use a better excitation map to replace the initial guidance. In this way, we design an additional excitation strategy in the testing phase to refine the predicted saliency map without any further training, we call it online excitation. Users can refine the saliency result by recurrently replacing the excitation maps with previous video saliency prediction outputs for better guidance. It provides an additional option for the users to trade-off between the saliency prediction quality and the computational cost during testing. Theoretically, if the excitation map is the same as the ground truth saliency map, our network can give the optimal solution of saliency prediction. We have conducted an experiment in the ablation study to prove the effectiveness of our online excitation. 
\section{Experiments}
\subsection{Implementation Details}\label{sec:del}
Our method is trained with three datasets DUTS~\cite{DUTS}, DAVIS~\cite{DAVIS} and DAVSOD~\cite{Fan_2019_CVPR}. Images are loaded into a batch according to their dataset, and we alternately train the Spatial Excitation branch with images from DUTS and DAVIS, Temporal Excitation branch with optical flow from DAVIS and DAVSOD, and the whole model with video from DAVIS and DAVSOD. The optimizer is SGD with momentum 0.9 and weight decay 0.0005. The learning rate starts from 5e-4 and decay to 1e-6. The curriculum rate initially set to 1, and decays following a cosine function.For data argumentation, all inputs are randomly flip horizontally and vertically. During testing every inputs will be resized to 256$\times$256. It takes about 40 hours to converge, which is one and a half times shorter (80 hours) than training without excitation. It shows that our excitation not only boosts the performance as shown below but also accelerates the training process.

\subsection{Datasets}
We conduct the experiments on four most frequently used VSOD datasets, including Freiburg-Berkeley motion segmentation dataset (FBMS)~\cite{FBMS}, video salient object detection dataset (ViSal)~\cite{ViSal}, densely annotation video segmentation dataset (DAVIS)~\cite{DAVIS}, and densely annotation video salient object detection dataset (DAVSOD)~\cite{Fan_2019_CVPR}.
\emph{\textbf{FBMS}} contains 59 videos with only 720 annotated frames. There are 29 videos for training and the rest of them are for testing.
\emph{\textbf{DAVIS}} is a high quality and high resolution densely annotated dataset under two resolutions, 480p and 1080p. There are 50 video sequences with 3455 densely annotated frames in pixel level. 
\emph{\textbf{ViSal}} is the first dataset specially designed for video salient object detection which includes 17 videos and 193 manual annotated frames.
\emph{\textbf{DAVSOD}} is the latest and most challenging video salient detection dataset with pixel-wise annotations and eye-fixation labels. We follow the same setting of SSAV\cite{Fan_2019_CVPR} and evaluate on 35 test videos.

\subsection{Evaluation Metrics}
We take three measurements to evaluate our methods: MAE~\cite{MAE}, F-measure~\cite{F_beta}, Structural measurement (S-measure)~\cite{Smeasure}. MAE is the mean absolute value between predicted saliency map and the groundtruth.

$F_\beta$ takes both precision and recall into consideration:
\begin{equation}
    F_\beta = \frac{(1+\beta^2)\times Precision\times Recall}{\beta^2\times Precision+ Recall},
\end{equation}
where $\beta^2$ is usually set to 0.3 and we use maximum $F_\beta$ for evaluation.

S-measure takes both region and object structural similarity into consideration:
\begin{equation}
    S = \mu * S_o + (1-\mu) * S_r,
\end{equation}
where $S_0$ and $S_r$ denotes the region-aware structural similarity and object-aware structural similarity respectively. $\mu$ is set to 0.5.

\subsection{Comparisons with State-of-the-arts}

\renewcommand{\arraystretch}{1.5} 
\begin{table*}[t]

  \centering
  \fontsize{8}{8}\selectfont
  \resizebox{\textwidth}{!}{
  \begin{threeparttable}

  \label{tab:performance_comparison}

    \begin{tabular}{cc|ccc|ccc|ccc|ccc}
    \toprule
    \multicolumn{2}{c}{\multirow{2}*{Method}}&
    \multicolumn{3}{c}{ FBMS}&\multicolumn{3}{c}{ ViSal}&\multicolumn{3}{c}{ DAVIS}&\multicolumn{3}{c}{ DAVSOD}\cr
    \cmidrule(lr){3-5} \cmidrule(lr){6-8} \cmidrule(lr){9-11} \cmidrule(lr){12-14}
    & &MAE$\downarrow$ &max $F_\beta$ $\uparrow$ &S $\uparrow$ &MAE$\downarrow$ &max $F_\beta$ $\uparrow$ &S $\uparrow$ &MAE$\downarrow$ &max $F_\beta$ $\uparrow$ &S $\uparrow$ &MAE$\downarrow$ &max $F_\beta$ $\uparrow$ &S $\uparrow$ \cr
    \midrule
    \rowcolor{mygray}

    DSS\cite{DSS}&I &0.080 &0.760 & 0.831&0.024 &0.917 &0.925& 0.059&0.720 &0.791& 0.112& 0.545& 0.630\cr

    BMPM\cite{BMPM}& I &0.056&0.791 &0.844 &0.022 &0.925& 0.930 &0.046&0.769 &0.834 &\blue{0.089}&0.599&0.704  \cr
    \rowcolor{mygray}
    BASNet\cite{BASNET}& I &0.051 &0.817&0.861 &\red{\textbf{0.011}} &\red{\textbf{0.949}} & \green{0.945} & 0.029&0.818&0.862 &0.110&0.597 &0.670\cr

    SIVM\cite{SIVM}& V&0.233 &0.416 &0.551 &0.199 &0.521 &0.611&0.211&0.461&0.551&0.291&0.299&0.491\cr
    \rowcolor{mygray}
    MSTM\cite{MSTM}& V&0.177&0.501&0.617&0.091 &0.681 &0.744 &0.166 &0.437 &0.588&0.210&0.341&0.529\cr

    SFLR\cite{SFLR}& V&0.119 &0.665&0.690&0.059 &0.782 &0.815 &0.055 &0.726 &0.781&0.132&0.477&0.627\cr

    \rowcolor{mygray}
    SCOM\cite{SCOM}&  V &0.078 &0.796 &0.789 &0.110 &0.829 &0.761&0.048&0.789&0.836&0.217&0.461&0.603\cr

    SCNN\cite{SCNN}&  V&0.091&0.766&0.799& 0.072&0.833 &0.850 &0.066 &0.711 &0.785&0.129&0.533&0.677\cr
    \rowcolor{mygray}
    FCNS\cite{VSOD_FCN}&  V&0.095 &0.745 &0.788 &0.045 &0.851 &0.879&0.055&0.711&0.781&0.121&0.545&0.664\cr

    FGRNE\cite{Flow_encoder}& V&0.085&0.771&0.811 &0.041 &0.850 &0.861 &0.043&0.782&0.840&0.099&0.577&0.701\cr
    \rowcolor{mygray}
    PDBM\cite{Pyramid}&  V&0.066 &0.801&0.845 &0.022 &0.916&0.929&0.028 &0.850&0.880&0.107&0.585&0.699\cr

    SSAV\cite{Fan_2019_CVPR}&  V &0.044 &0.855 &0.873 &0.018 &0.939 &0.943&0.029 &0.861 &0.891&0.092&0.602&0.719\cr
	
    \rowcolor{mygray}
    MGAN\cite{motion}&  V &\blue{0.028} &\green{0.889} &\blue{0.907} &\blue{0.015}&\blue{0.944} &\blue{0.944} &\blue{0.022} &\green{0.897} &\green{0.911}&0.114&\blue{0.627}&\blue{0.737}\cr

    Ours& V&\green{0.027}&\blue{0.887}&\green{0.910} &\green{0.014}&\green{0.947}& 0.943&\green{0.021}&\blue{0.894} &\blue{0.905} &\green{0.078}&\green{0.648}&\green{0.753}\cr

	\rowcolor{mygray}
    Ours$^{\star}$& V&\red{\textbf{0.026}}& \red{\textbf{0.897}}&\red{\textbf{0.915}} &\green{ 0.014}&\red{\textbf{ 0.949}}&\red{\textbf{ 0.946}}&\red{\textbf{0.019}} &\red{\textbf{0.904}} &\red{\textbf{0.916}}& \red{\textbf{0.074}} &\red{\textbf{0.664}} &\red{\textbf{0.780}}\cr
    \bottomrule
    \end{tabular}
    \caption{Quantitative comparison with image salient object detection methods (labeled as I) and the state-of-the-art VSOD methods (labelled as V) by three evaluation metrics. $Ours$ and $Ours^\star$ indicate the results without and with the online excitation. Top three performances are marked in \red{Red}, \green{Green}, and \blue{Blue} respectively.
    }
    \label{Results}
    \end{threeparttable}}
\end{table*}

\begin{figure*}[t]
\centering
\hspace{-5mm}
\subfloat[frame]{
\begin{minipage}[b]{0.1\textwidth}
\centering
\includegraphics[scale=0.1]{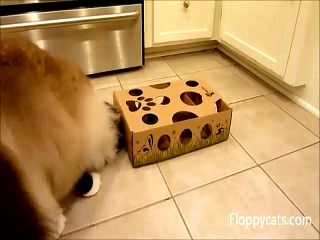} \\
\includegraphics[scale=0.1]{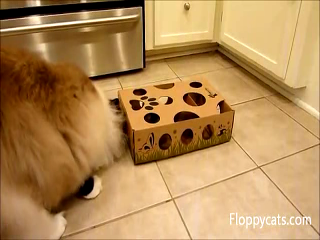} \\
\includegraphics[scale=0.1]{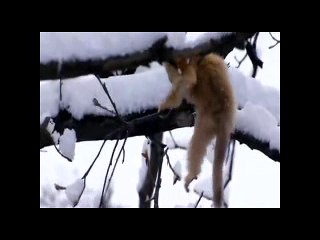} \\
\includegraphics[scale=0.1]{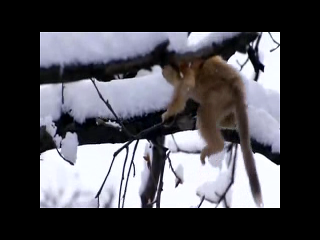} \\
\includegraphics[scale=0.1]{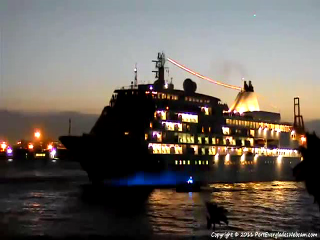} \\
\includegraphics[scale=0.1]{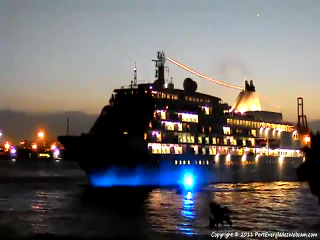} \\
\includegraphics[scale=0.1]{./figs/fig2/image11.png} \\
\includegraphics[scale=0.1]{./figs/fig2/image12.png} 
\end{minipage}
}\hspace{-2.5mm}
\subfloat[GT]{
\begin{minipage}[b]{0.1\textwidth}
\centering
\includegraphics[scale=0.1]{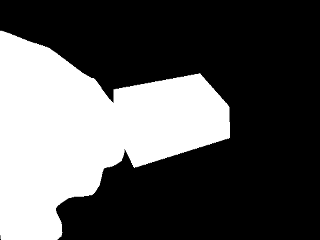} \\
\includegraphics[scale=0.1]{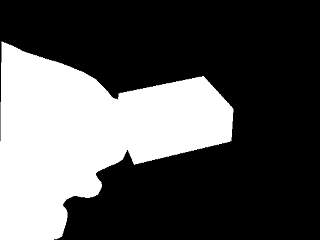} \\
\includegraphics[scale=0.1]{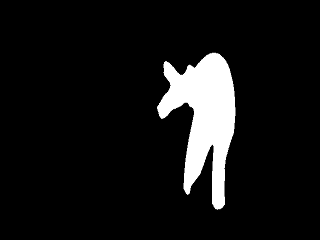} \\
\includegraphics[scale=0.1]{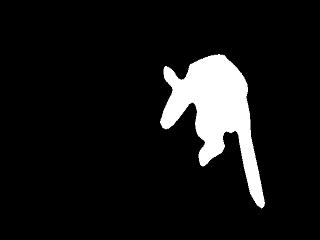} \\
\includegraphics[scale=0.1]{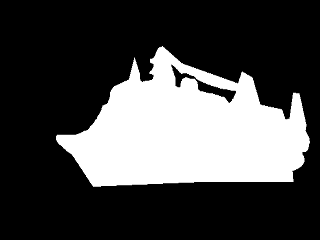} \\
\includegraphics[scale=0.1]{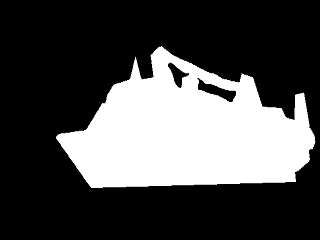} \\
\includegraphics[scale=0.1]{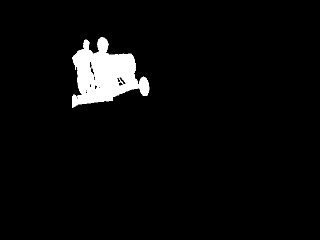} \\
\includegraphics[scale=0.1]{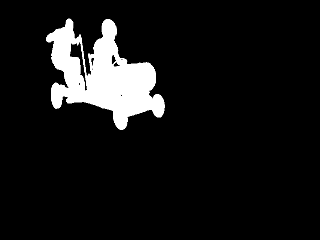}
\end{minipage}
}\hspace{-2.5mm}
\subfloat[Ours]{
\begin{minipage}[b]{0.1\textwidth}
\centering
\includegraphics[scale=0.1]{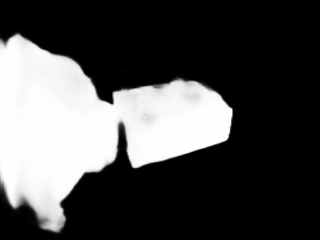}\\
\includegraphics[scale=0.1]{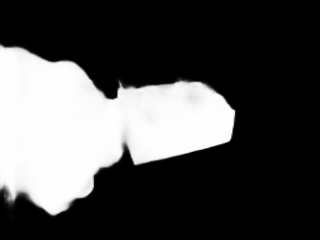}\\
\includegraphics[scale=0.1335]{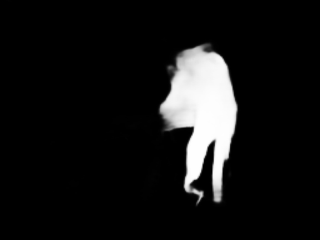}\\
\includegraphics[scale=0.1335]{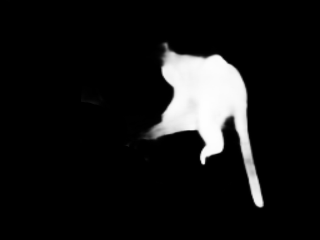}\\
\includegraphics[scale=0.1]{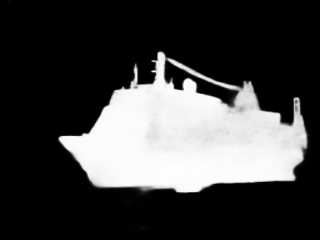}\\
\includegraphics[scale=0.1]{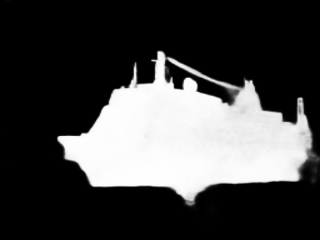}\\
\includegraphics[scale=0.1]{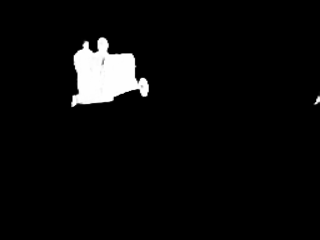}\\
\includegraphics[scale=0.1]{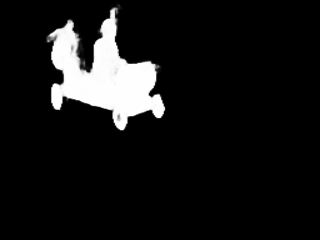}
\end{minipage}
}\hspace{-2.5mm}
\subfloat[MGAN]{
	\begin{minipage}[b]{0.1\textwidth}
		\centering
		\includegraphics[scale=0.1]{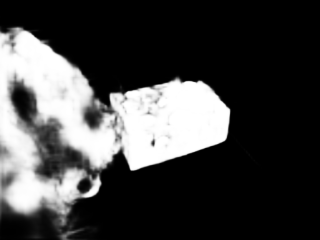} \\
		\includegraphics[scale=0.1]{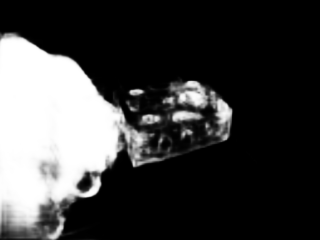} \\
		\includegraphics[scale=0.1]{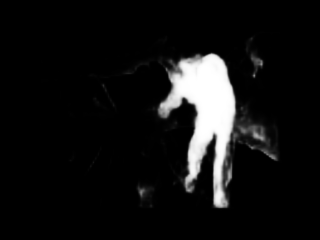} \\
		\includegraphics[scale=0.1]{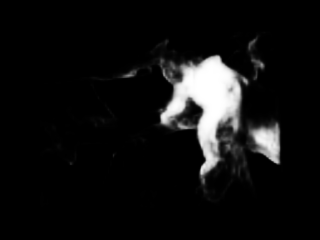} \\
		\includegraphics[scale=0.1]{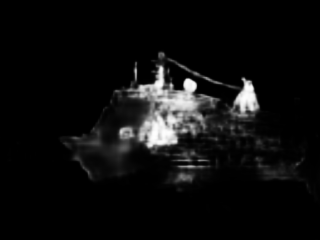} \\
		\includegraphics[scale=0.1]{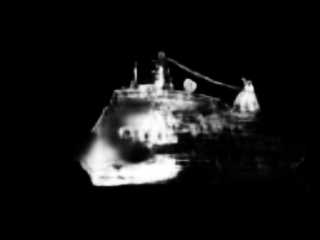}\\
		\includegraphics[scale=0.1]{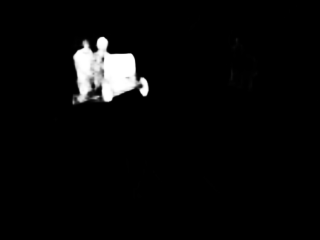}\\
		\includegraphics[scale=0.1]{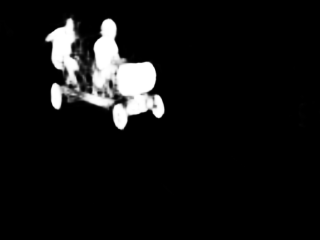}
	\end{minipage}
}\hspace{-2.5mm}
\subfloat[SSAV]{
\begin{minipage}[b]{0.1\textwidth}
\centering
\includegraphics[scale=0.1]{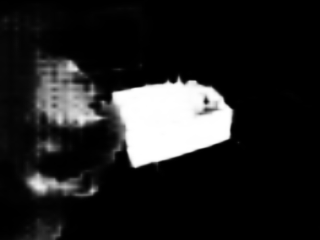} \\
\includegraphics[scale=0.1]{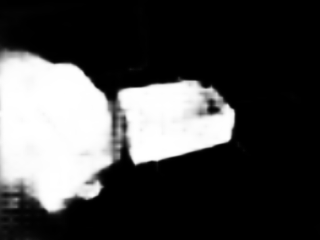} \\
\includegraphics[scale=0.1]{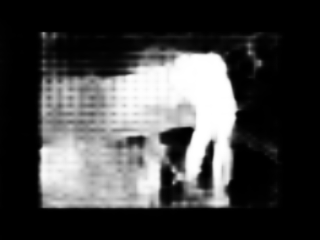} \\
\includegraphics[scale=0.1]{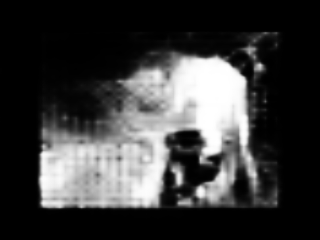} \\
\includegraphics[scale=0.1]{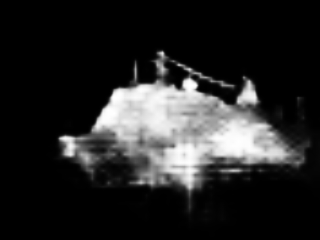} \\
\includegraphics[scale=0.1]{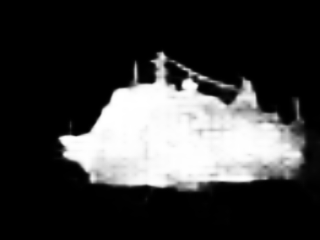}\\
\includegraphics[scale=0.1]{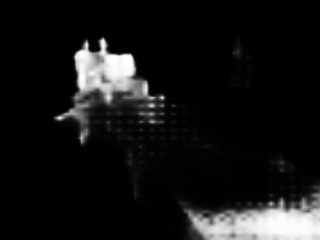} \\
\includegraphics[scale=0.1]{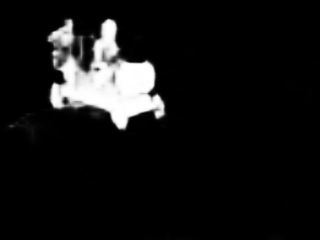}
\end{minipage}
}\hspace{-2.5mm}
\subfloat[PDBM]{
\begin{minipage}[b]{0.1\textwidth}
\centering
\includegraphics[scale=0.1]{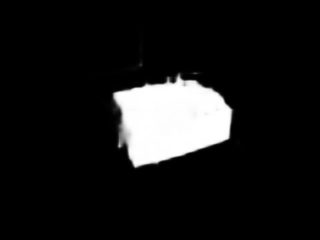} \\
\includegraphics[scale=0.1]{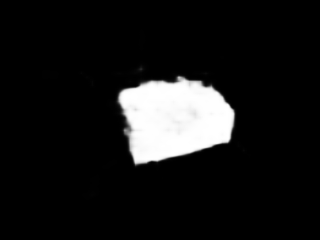} \\
\includegraphics[scale=0.1]{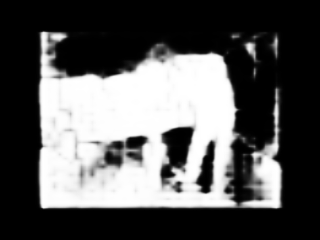} \\
\includegraphics[scale=0.1]{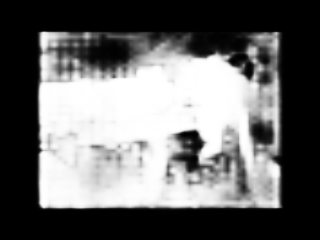} \\
\includegraphics[scale=0.1]{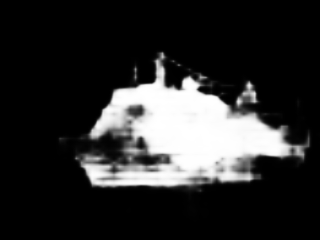} \\
\includegraphics[scale=0.1]{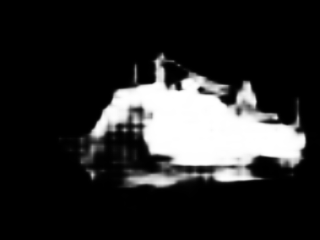} \\
\includegraphics[scale=0.1]{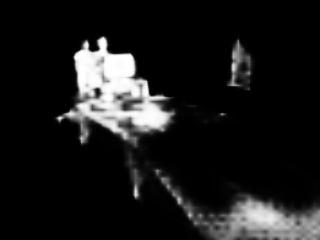} \\
\includegraphics[scale=0.1]{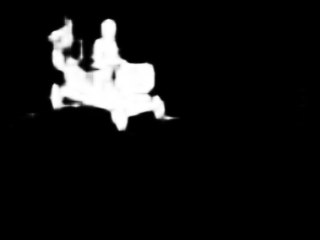} 
\end{minipage}
}\hspace{-2.5mm}
\subfloat[FCNS]{
\begin{minipage}[b]{0.1\textwidth}
\centering
\includegraphics[scale=0.1]{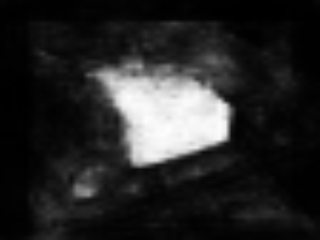} \\
\includegraphics[scale=0.1]{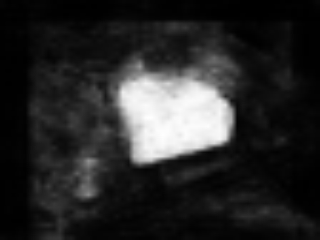} \\
\includegraphics[scale=0.1]{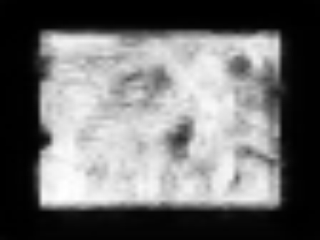} \\
\includegraphics[scale=0.1]{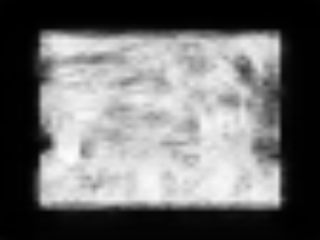} \\
\includegraphics[scale=0.1]{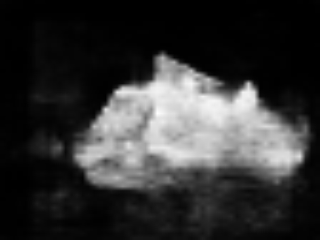} \\
\includegraphics[scale=0.1]{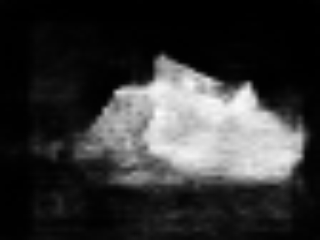} \\
\includegraphics[scale=0.1]{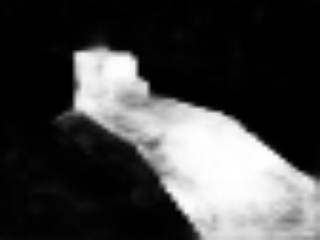} \\
\includegraphics[scale=0.1]{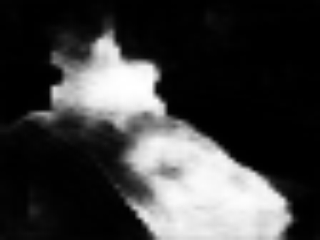} 
\end{minipage}
}\hspace{-2.5mm}
\subfloat[BASNet]{
\begin{minipage}[b]{0.1\textwidth}
\centering
\includegraphics[scale=0.1]{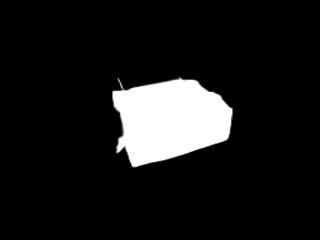} \\
\includegraphics[scale=0.1]{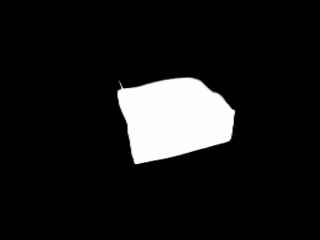} \\
\includegraphics[scale=0.1]{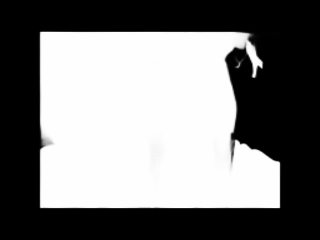} \\
\includegraphics[scale=0.1]{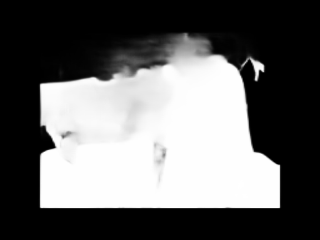} \\
\includegraphics[scale=0.1]{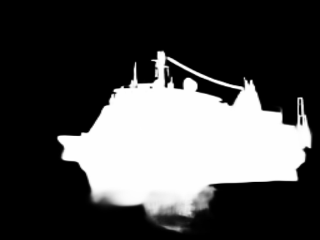} \\
\includegraphics[scale=0.1]{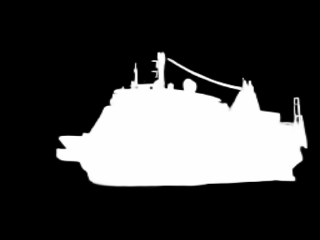} \\
\includegraphics[scale=0.1]{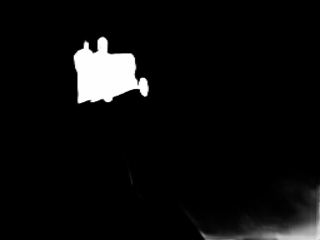} \\
\includegraphics[scale=0.1]{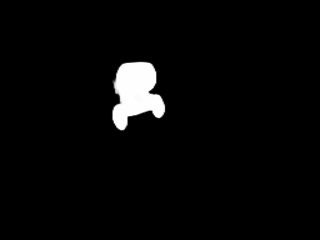} 
\end{minipage}
}\hspace{-2.5mm}
\subfloat[BMPM]{
\begin{minipage}[b]{0.1\textwidth}
\centering
\includegraphics[scale=0.1]{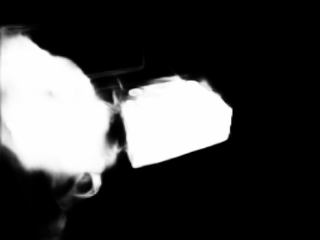} \\
\includegraphics[scale=0.1]{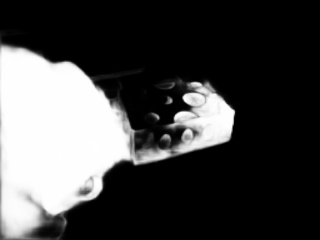} \\
\includegraphics[scale=0.1]{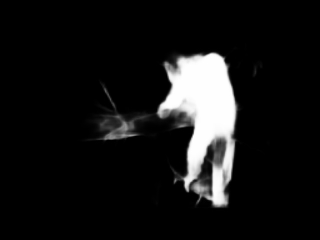} \\
\includegraphics[scale=0.1]{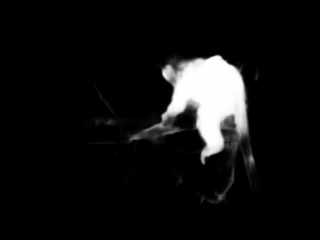} \\
\includegraphics[scale=0.1]{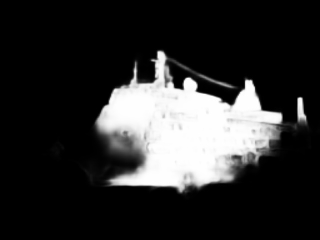} \\
\includegraphics[scale=0.1]{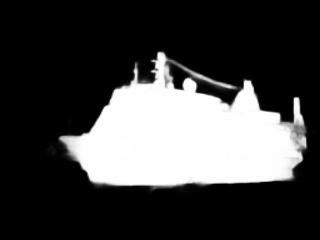} \\
\includegraphics[scale=0.1]{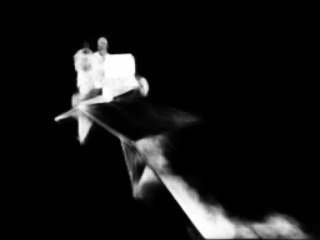} \\
\includegraphics[scale=0.1]{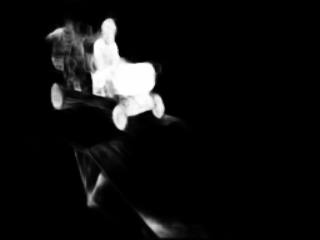} 
\end{minipage}
}\hspace{-2.5mm}
\subfloat[DSS]{
\begin{minipage}[b]{0.1\textwidth}
\centering
\includegraphics[scale=0.1]{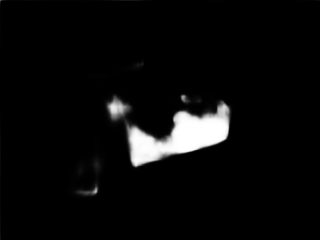} \\
\includegraphics[scale=0.1]{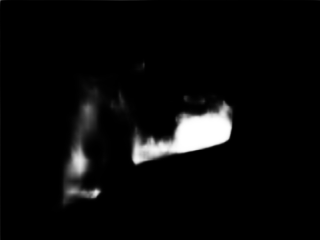} \\
\includegraphics[scale=0.1]{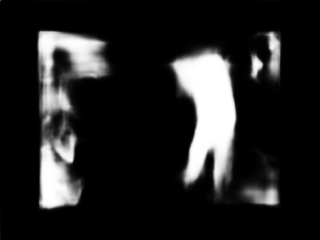} \\
\includegraphics[scale=0.1]{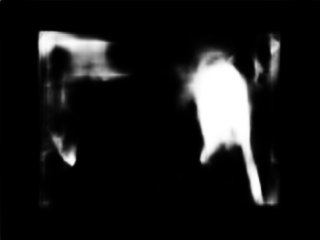} \\
\includegraphics[scale=0.1]{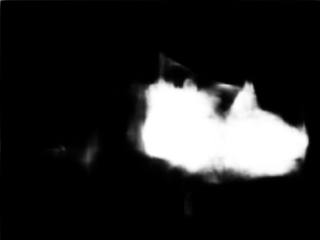} \\
\includegraphics[scale=0.1]{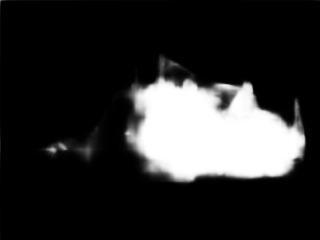} \\
\includegraphics[scale=0.1]{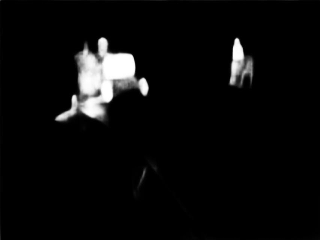} \\
\includegraphics[scale=0.1]{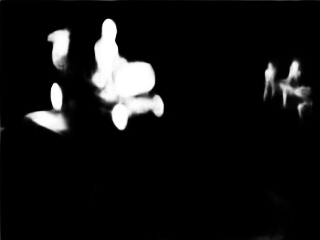} 
\end{minipage}
}\hspace{-5mm}
\caption{Qualitative comparison with state-of-the-art methods. Our TENet produces clear object boundaries while capture temporally salient objects in the video.}
\label{img:Quantitative}
\end{figure*}
We compare our method with 13 saliency methods, including image salient object detection methods (DSS~\cite{DSS}, BMPM~\cite{BMPM}, BASNet~\cite{BASNET}) and video salient object detection methods (SIVM~\cite{SIVM}, MSTM~\cite{MSTM}, SFLR~\cite{SFLR}, SCOM~\cite{SCOM}, SCNN~\cite{SCNN}, FCNS~\cite{VSOD_FCN}, FGRNE~\cite{Flow_encoder}, PDBM~\cite{Pyramid}, SSAV~\cite{Fan_2019_CVPR}, MGAN~\cite{motion}).

Table~\ref{Results} show the quantitative evaluation with existing methods. For image salient object detection methods, they perform well in some video salient datasets, because objects that distinct in appearance draws the most attention from the viewer if they are not moving dramatically in the video. On the other hand, although the most recent VSOD method SSAV~\cite{Fan_2019_CVPR} leverages eye-fixation information to guide the network, the imbalance of spatial and temporal domains harm the accuracy of their saliency results. Our method performs the best statistics results among all the methods in all datasets. Note that,  `$Ours$' is the model with the excitation map generated by the excitation prediction branch, \ie, without online excitation. `$Ours^\star$' indicates the results involve online excitation by recurrently applying the previous network outputs. We can find a significant improvement when we apply online excitation. It proves that a precise excitation map can give more accurate guidance for the network, even without further training.

Another interesting observation that can be found in Table~\ref{Results} is that the datasets also affect the network performance a lot. Because when looking into a video, people tend to focus on moving objects. Some moving objects are not salient in the single frame will definitely distract the network. For some easy dataset whose salient objects are moving and occupy a large part of the image, like~\emph{ViSal}, the performance is good for both image-based methods and video-based methods. While in some complicated datasets like~\emph{DAVSOD} and~\emph{FBMS}, the salient objects in the temporal domain are unfortunately not salient in the spatial domain. The statistical results of them are much worse than those easier datasets. Since our proposed excitation mechanism governs both the spatial and temporal information, our methods gain a much higher performance than the existing methods.

Fig.~\ref{img:Quantitative} shows the qualitative comparison. The results predicted by image-based methods shown in Fig.~\ref{img:Quantitative} (h)-(j) fails to detect the object region accurately and sometimes cannot distinguish the foreground and background due to the lack of temporal information. VSOD methods shown in Fig.~\ref{img:Quantitative} (d)-(g) provide visually more reasonable saliency maps. However, the boundary of the salient object is not clear and the inside region is blurry, due to the contradictory spatial and temporal features. In contrast, our results (without online excitation, see Fig.~\ref{img:Quantitative}(c)) show clear boundaries as well as high-confidence interior salient regions. 

\subsection{Ablation Study}
In this section, we explore the effectiveness of our proposed modules. We test the performance on DAVSOD which is the most challenging VSOD dataset.
\subsubsection{Effectiveness of Triple Excitation.}
\renewcommand\arraystretch{1.2}
\begin{table}[t]
\center
\small
\resizebox{\textwidth}{!}{
\begin{tabular}{>{\centering}p{0.4cm}|>{\centering}p{0.60cm}|>{\centering \columncolor{mygray}}p{0.70cm}|>{\centering \columncolor{mygray}}p{0.70cm}|>{\centering \columncolor{mygray}}p{0.70cm}|>{\centering \columncolor{mygray}}p{0.70cm}|>{\centering}p{0.70cm}|>{\centering}p{0.70cm}|>{\centering}p{0.70cm}|>{\centering}p{0.70cm}|>{\centering \columncolor{mygray}}p{0.70cm}|>{\centering \columncolor{mygray}}p{0.70cm}|>{\centering \columncolor{mygray}}p{0.70cm}|>{\centering \columncolor{mygray}}p{0.70cm}}
\toprule
\multicolumn{2}{ c|}{Temporal} &\checkmark&\checkmark&\checkmark &\checkmark&&&&&\checkmark&\checkmark&\checkmark&\checkmark \cr
\multicolumn{2}{ c|}{Spatial} &&&&&\checkmark&\checkmark&\checkmark&\checkmark&\checkmark&\checkmark&\checkmark&\checkmark \cr
\hline
\multicolumn{2}{ c|}{Online (1)} &&\checkmark&&&&\checkmark&&&&\checkmark&&\cr
\multicolumn{2}{ c|}{Online (20)} &&&\checkmark&&&&\checkmark&&&&\checkmark&\cr
\multicolumn{2}{ c|}{Online (GT)} &&&&\checkmark&&&&\checkmark&&&&\checkmark\cr
\hline
\hline
\multirow{3}*{\rotatebox{90}{Results}}&MAE &0.092  &0.090  &0.091&0.069&0.084&0.081&0.080&0.062&0.078&0.075&0.074& 0.053\cr
&$F_{\beta}$ &0.591  &0.595  &0.594&0.691&0.615&0.628&0.631&0.688&0.648&0.659&0.664&0.841\cr
&S &0.693  &0.702  &0.708&0.738&0.715&0.733&0.741&0.764&0.753&0.772&0.780&0.862 \cr
\bottomrule
\end{tabular}}
\caption{Ablation study for triple excitation mechanism on the DAVSOD dataset. We separately demonstrate the effectiveness of each excitation component. Online (N) indicates the online excitation with N iterations.}
\label{tab:Ablation}
\end{table}
Table~\ref{tab:Ablation} shows an ablation study to evaluate the effectiveness of our triple excitation method. In this experiment, we choose 14 configurations of different excitation strategies. The checkmark in Table~\ref{tab:Ablation} indicates the activated excitation component. We can observe that both temporal and spatial excitations boost the detection performance by a large margin (comparing to the first column without checkmark).

We then demonstrate the proposed online excitation. We perform online excitation for one and multiple iterations, labeled as $Online (1)$ and $Online (20)$. In our experiment, applying more than 20 iterations cannot bring extra improvement. We also show the ideal case, in which uses the ground truth saliency map as the excitation map, labeled as $Online (GT)$ in Table~\ref{tab:Ablation}. Although using ground truth as excitation information is impossible in testing, we demonstrate the upper-bound of our method. It proves that a more accuracy excitation map will bring more profits which also implicitly demonstrates the effectiveness of the excitation mechanism.

The statistical results reveal that our three excitations all together work well as we expected. Online excitation does not only introduces more precise excitation guidance but also provides additional optional to exchange the computational cost for the prediction accuracy, by iteratively running the online excitation. $Online (20)$ shows the results of 20 times online iterations. We can find an obvious improvement in three measurements. `$Ours^{\star}$' in Table~\ref{Results} is implemented with 20 iterations.

\subsubsection{Effectiveness of Semi-curriculum Learning.} Our semi-curriculum learning involves two main components, GT and learned complementary maps. We consider using the GT only as the traditional curriculum solution. As can be seen in Table \ref{fig:curriculum}, the traditional curriculum learning hugely reduces the convergence time. However, as the network relies too much on the perfect ground truth, once no guidance is provided during testing, the curriculum learning strategy does not bring too much improvement. On the other hand, using a learned complementary map for excitation can ease this problem. To provide initial supervision of the network, we pretrain the complementary map for static saliency detection. Using the learned complementary map can provide guidance during testing, which is the key to maintain consistent performances between training and testing phrases. The main limitation is that it requires a separated pretraining, which largely increases the convergence time. The proposed semi-curriculum learning strategy remedies the limitations of previous two methods, leading to faster and better convergence.
\renewcommand{\arraystretch}{1}
\begin{table}[t]
	\centering
		\begin{threeparttable}
			\label{tab:performance_comparison}
			
			\begin{tabular}{p{5cm}|ccc|cc}
				\toprule
				\multirow{2}*{Method} &\multicolumn{4}{c}{ DAVSOD} \cr
				\cmidrule(lr){2-5}
				&MAE$\downarrow$ &max $F_\beta$ $\uparrow$ &S $\uparrow$ & Convergence Time \cr
				\midrule
				Baseline& 0.112&0.579 &0.694 & 80 hours\cr
				Baseline + Curriculum& 0.108&0.584 &0.699 & 32 hours \cr
				Baseline + Learned Excitation& 0.080&0.641 &0.743 & 46 (pre-training) + 38 hours\cr
                \midrule
				Baseline + Semi-curriculum& 0.078&0.648 &0.753 & 40 hours \cr
				\bottomrule
			\end{tabular}
	\end{threeparttable}
	\caption{Ablation study on the proposed semi-curriculum learning.}
	\label{fig:curriculum}
\end{table}

\subsection{Timing Statistics}
We also show the running time of different models in Table~\ref{tab:running time}. All the methods are tested on the same platform: Intel(R) Xeon(R) CPU E5-2620v4 @2.10GHz and GTX1080Ti. The timing statistics do not include the pre-/post-processing time. Due to our plug-and-play excitations, we can have a fast timing performance compared with most of the deep learning based VSOD methods.

\begin{table}[t]
    \centering
    \setlength{\tabcolsep}{1.2mm}
    \begin{tabular}{>{\centering}p{1.7cm}|>{\centering}p{1.7cm}>{\centering}p{1.7cm}>{\centering}p{1.7cm}>{\centering}p{1.7cm}>{\centering}p{1.7cm}c}
    \hline
         Method&SIVM\cite{SIVM}&\hspace{-0.2cm}BMPM\cite{BMPM}&\hspace{-0.2cm}FCNS\cite{VSOD_FCN} &\hspace{-0.2cm}PMDB\cite{Pyramid}&\hspace{-0.2cm} SSAV\cite{Fan_2019_CVPR} &\hspace{-0.2cm}Ours  \\
         \hline
         Time(s)&18.1&\hspace{-0.2cm}0.03&\hspace{-0.2cm}0.50&\hspace{-0.2cm}0.08&\hspace{-0.2cm}0.08&\hspace{-0.2cm}0.06  \\
         \hline
    \end{tabular}
    \caption{Running time comparison of existing methods.}
    \label{tab:running time}
\end{table}

\section{Conclusion}
This paper proposes a novel video salient object detection method equipped with a triple excitation mechanism. Spatial and temporal excitations are proposed during training phase to tackle the saliency shifting and contradictory spatio-temporal features problems. Besides, we introduce semi-curriculum learning during training to loosen the task difficulty at first and reach a better converage. Furthermore, we propose the first online excitation in testing phase to allow the network keep refining the saliency result by using the network output saliency map for excitation. Extensive experiments show that our results outperform all the competitors.

\section*{Acknowledgement}
This project is supported by the National Natural Science Foundation of China (No. 61472145, No. 61972162, and No. 61702194), the Special Fund of Science and Technology Research and Development of Applications From Guangdong Province (SF-STRDA-GD) (No. 2016B010127003), the Guangzhou Key Industrial Technology Research fund (No. 201802010036), the Guangdong Natural Science Foundation (No. 2017A030312008), and the CCF-Tencent Open Research fund (CCF-Tencent RAGR20190112).
%
%
\bibliographystyle{splncs04}
\bibliography{egbib}

\begin{thebibliography}{10}
\providecommand{\url}[1]{\texttt{#1}}
\providecommand{\urlprefix}{URL }
\providecommand{\doi}[1]{https://doi.org/#1}

\bibitem{F_beta}
Achanta, R., Hemami, S., Estrada, F., S{\"u}sstrunk, S.: Frequency-tuned
  salient region detection. In: CVPR. pp. 1597--1604 (2009)

\bibitem{Bengio:2009}
Bengio, Y., Louradour, J., Collobert, R., Weston, J.: Curriculum learning. In:
  ICML. pp. 41--48 (2009)

\bibitem{top-bottom}
Borji, A.: Boosting bottom-up and top-down visual features for saliency
  estimation. In: CVPR. pp. 438--445. IEEE (2012)

\bibitem{FBMS}
Brox, T., Malik, J.: Object segmentation by long term analysis of point
  trajectories. In: ECCV. pp. 282--295 (2010)

\bibitem{seg_detect3}
Cao, J., Pang, Y., Li, X.: Triply supervised decoder networks for joint
  detection and segmentation. In: CVPR. pp. 7392--7401 (2019)

\bibitem{SFLR}
Chen, C., Li, S., Wang, Y., Qin, H., Hao, A.: Video saliency detection via
  spatial-temporal fusion and low-rank coherency diffusion. IEEE TIP
  \textbf{26}(7),  3156--3170 (2017)

\bibitem{SCOM}
Chen, Y., Zou, W., Tang, Y., Li, X., Xu, C., Komodakis, N.: Scom:
  Spatiotemporal constrained optimization for salient object detection. IEEE
  TIP  \textbf{27}(7),  3345--3357 (2018)

\bibitem{seg_detect1}
Dai, J., He, K., Sun, J.: Instance-aware semantic segmentation via multi-task
  network cascades. In: CVPR. pp. 3150--3158 (2016)

\bibitem{BCE}
De~Boer, P.T., Kroese, D.P., Mannor, S., Rubinstein, R.Y.: A tutorial on the
  cross-entropy method. Annals of operations research  \textbf{134}(1),  19--67
  (2005)

\bibitem{Smeasure}
Fan, D.P., Cheng, M.M., Liu, Y., Li, T., Borji, A.: Structure-measure: A new
  way to evaluate foreground maps. In: ICCV. pp. 4548--4557 (2017)

\bibitem{Fan_2019_CVPR}
Fan, D.P., Wang, W., Cheng, M.M., Shen, J.: Shifting more attention to video
  salient object detection. In: CVPR. pp. 8554--8564 (2019)

\bibitem{boundaryloss}
Feng, M., Lu, H., Ding, E.: Attentive feedback network for boundary-aware
  salient object detection. In: CVPR (2019)

\bibitem{bottom1}
Gao, D., Vasconcelos, N.: Bottom-up saliency is a discriminant process. In:
  ICCV. pp.~1--6 (2007)

\bibitem{seg_detect2}
Hariharan, B., Arbel{\'a}ez, P., Girshick, R., Malik, J.: Simultaneous
  detection and segmentation. In: ECCV. pp. 297--312 (2014)

\bibitem{detect1}
He, K., Gkioxari, G., Doll{\'a}r, P., Girshick, R.: Mask r-cnn. In: ICCV. pp.
  2961--2969 (2017)

\bibitem{ResNet}
He, K., Zhang, X., Ren, S., Sun, J.: Deep residual learning for image
  recognition. In: CVPR. pp. 770--778 (2016)

\bibitem{DSS}
Hou, Q., Cheng, M.M., Hu, X., Borji, A., Tu, Z., Torr, P.H.: Deeply supervised
  salient object detection with short connections. In: CVPR. pp. 3203--3212
  (2017)

\bibitem{attention_shift2}
Itti, L., Koch, C., Niebur, E.: A model of saliency-based visual attention for
  rapid scene analysis. IEEE TPAMI (11),  1254--1259 (1998)

\bibitem{bottom2}
Jiang, H., Wang, J., Yuan, Z., Wu, Y., Zheng, N., Li, S.: Salient object
  detection: A discriminative regional feature integration approach. In: CVPR.
  pp. 2083--2090 (2013)

\bibitem{eyefixation1}
Jiang, M., Huang, S., Duan, J., Zhao, Q.: Salicon: Saliency in context. In:
  CVPR (June 2015)

\bibitem{attention_shift3}
Koch, C., Ullman, S.: Shifts in selective visual attention: towards the
  underlying neural circuitry. In: Matters of intelligence, pp. 115--141.
  Springer (1987)

\bibitem{object_tracking1}
Lee, H., Kim, D.: Salient region-based online object tracking. In: WACV. pp.
  1170--1177. IEEE (2018)

\bibitem{Flow_encoder}
Li, G., Xie, Y., Wei, T., Wang, K., Lin, L.: Flow guided recurrent neural
  encoder for video salient object detection. In: ICCV. pp. 3243--3252 (2018)

\bibitem{saliency_classification1}
Li, G., Yu, Y.: Visual saliency based on multiscale deep features. In: CVPR.
  pp. 5455--5463 (2015)

\bibitem{motion}
Li, H., Chen, G., Li, G., Yu, Y.: Motion guided attention for video salient
  object detection. In: ICCV (2019)

\bibitem{graph_cut2}
Li, S., Seybold, B., Vorobyov, A., Lei, X., Jay~Kuo, C.C.: Unsupervised video
  object segmentation with motion-based bilateral networks. In: ECCV. pp.
  207--223 (2018)

\bibitem{SOD2}
Liu, J.J., Hou, Q., Cheng, M.M., Feng, J., Jiang, J.: A simple pooling-based
  design for real-time salient object detection. In: CVPR (2019)

\bibitem{Attention1}
Liu, N., Han, J., Yang, M.H.: Picanet: Learning pixel-wise contextual attention
  for saliency detection. In: CVPR. pp. 3089--3098 (2018)

\bibitem{liu2019selflow}
Liu, P., Lyu, M., King, I., Xu, J.: Selflow: Self-supervised learning of
  optical flow. In: CVPR. pp. 4571--4580 (2019)

\bibitem{long2015fully}
Long, J., Shelhamer, E., Darrell, T.: Fully convolutional networks for semantic
  segmentation. In: CVPR. pp. 3431--3440 (2015)

\bibitem{Lu_2019_CVPR}
Lu, X., Wang, W., Ma, C., Shen, J., Shao, L., Porikli, F.: See more, know more:
  Unsupervised video object segmentation with co-attention siamese networks.
  In: CVPR (2019)

\bibitem{manipulation1}
Mechrez, R., Shechtman, E., Zelnik-Manor, L.: Saliency driven image
  manipulation. Machine Vision and Applications  \textbf{30}(2),  189--202
  (2019)

\bibitem{graph_cut1}
Papazoglou, A., Ferrari, V.: Fast object segmentation in unconstrained video.
  In: ICCV. pp. 1777--1784 (2013)

\bibitem{MAE}
Perazzi, F., Kr{\"a}henb{\"u}hl, P., Pritch, Y., Hornung, A.: Saliency filters:
  Contrast based filtering for salient region detection. In: CVPR. pp. 733--740
  (2012)

\bibitem{DAVIS}
Perazzi, F., Pont-Tuset, J., McWilliams, B., Van~Gool, L., Gross, M.,
  Sorkine-Hornung, A.: A benchmark dataset and evaluation methodology for video
  object segmentation. In: CVPR. pp. 724--732 (2016)

\bibitem{BASNET}
Qin, X., Zhang, Z., Huang, C., Gao, C., Dehghan, M., Jagersand, M.: Basnet:
  Boundary-aware salient object detection. In: CVPR. pp. 7479--7489 (2019)

\bibitem{SIVM}
Rahtu, E., Kannala, J., Salo, M., Heikkil{\"a}, J.: Segmenting salient objects
  from images and videos. In: ECCV. pp. 366--379 (2010)

\bibitem{manipulation2}
Shafieyan, F., Karimi, N., Mirmahboub, B., Samavi, S., Shirani, S.: Image seam
  carving using depth assisted saliency map. In: ICIP. pp. 1155--1159. IEEE
  (2014)

\bibitem{NIPS2015_5955}
SHI, X., Chen, Z., Wang, H., Yeung, D.Y., Wong, W.k., WOO, W.c.: Convolutional
  lstm network: A machine learning approach for precipitation nowcasting. In:
  NeurIPS. pp. 802--810 (2015)

\bibitem{Pyramid}
Song, H., Wang, W., Zhao, S., Shen, J., Lam, K.M.: Pyramid dilated deeper
  convlstm for video salient object detection. In: ECCV. pp. 715--731 (2018)

\bibitem{attention_shift1}
Squire, L.R., Dronkers, N., Baldo, J.: Encyclopedia of neuroscience. Elsevier
  (2009)

\bibitem{SCNN}
Tang, Y., Zou, W., Jin, Z., Chen, Y., Hua, Y., Li, X.: Weakly supervised
  salient object detection with spatiotemporal cascade neural networks. IEEE
  Transactions on Circuits and Systems for Video Technology  (2018)

\bibitem{MSTM}
Tu, W.C., He, S., Yang, Q., Chien, S.Y.: Real-time salient object detection
  with a minimum spanning tree. In: CVPR. pp. 2334--2342 (2016)

\bibitem{video_understanding1}
Wang, H., Kl{\"a}ser, A., Schmid, C., Liu, C.L.: Dense trajectories and motion
  boundary descriptors for action recognition. IJCV  \textbf{103}(1),  60--79
  (2013)

\bibitem{video_understanding2}
Wang, H., Schmid, C.: Action recognition with improved trajectories. In: ICCV.
  pp. 3551--3558 (2013)

\bibitem{DUTS}
Wang, L., Lu, H., Wang, Y., Feng, M., Wang, D., Yin, B., Ruan, X.: Learning to
  detect salient objects with image-level supervision. In: CVPR. pp. 136--145
  (2017)

\bibitem{video_understanding3}
Wang, L., Qiao, Y., Tang, X.: Action recognition with trajectory-pooled
  deep-convolutional descriptors. In: CVPR. pp. 4305--4314 (2015)

\bibitem{eyefixation2}
Wang, W., Shen, J., Guo, F., Cheng, M.M., Borji, A.: Revisiting video saliency:
  A large-scale benchmark and a new model. In: CVPR (2018)

\bibitem{ViSal}
Wang, W., Shen, J., Shao, L.: Consistent video saliency using local gradient
  flow optimization and global refinement. IEEE TIP  \textbf{24}(11),
  4185--4196 (2015)

\bibitem{VSOD_FCN}
Wang, W., Shen, J., Shao, L.: Video salient object detection via fully
  convolutional networks. IEEE TIP  \textbf{27}(1),  38--49 (2017)

\bibitem{Wang_2019_CVPR}
Wang, W., Song, H., Zhao, S., Shen, J., Zhao, S., Hoi, S.C.H., Ling, H.:
  Learning unsupervised video object segmentation through visual attention. In:
  CVPR (June 2019)

\bibitem{SSIM}
Wang, Z., Bovik, A.C., Sheikh, H.R., Simoncelli, E.P., et~al.: Image quality
  assessment: from error visibility to structural similarity. IEEE TIP
  \textbf{13}(4),  600--612 (2004)

\bibitem{top}
Yang, J., Yang, M.H.: Top-down visual saliency via joint crf and dictionary
  learning. IEEE TPAMI  \textbf{39}(3),  576--588 (2016)

\bibitem{people_reid1}
Yang, Y., Yang, J., Yan, J., Liao, S., Yi, D., Li, S.Z.: Salient color names
  for person re-identification. In: ECCV. pp. 536--551. Springer (2014)

\bibitem{yang2019anchor}
Yang, Z., Wang, Q., Bertinetto, L., Bai, S., Hu, W., Torr, P.H.: Anchor
  diffusion for unsupervised video object segmentation. In: ICCV (2019)

\bibitem{iou}
Yu, J., Jiang, Y., Wang, Z., Cao, Z., Huang, T.: Unitbox: An advanced object
  detection network. In: ACM MM. pp. 516--520 (2016)

\bibitem{BMPM}
Zhang, L., Dai, J., Lu, H., He, Y., Wang, G.: A bi-directional message passing
  model for salient object detection. In: CVPR. pp. 1741--1750 (2018)

\bibitem{SOD1}
Zhang, X., Wang, T., Qi, J., Lu, H., Wang, G.: Progressive attention guided
  recurrent network for salient object detection. In: CVPR (June 2018)

\bibitem{detect2}
Zhang, Z., Qiao, S., Xie, C., Shen, W., Wang, B., Yuille, A.L.: Single-shot
  object detection with enriched semantics. In: CVPR. pp. 5813--5821 (2018)

\bibitem{saliency_classification2}
Zhao, R., Ouyang, W., Li, H., Wang, X.: Saliency detection by multi-context
  deep learning. In: CVPR. pp. 1265--1274 (2015)

\bibitem{people_reid3}
Zhao, R., Ouyang, W., Wang, X.: Unsupervised salience learning for person
  re-identification. In: CVPR. pp. 3586--3593 (2013)

\bibitem{people_reid2}
Zhao, R., Oyang, W., Wang, X.: Person re-identification by saliency learning.
  IEEE TPAMI  \textbf{39}(2),  356--370 (2016)

\bibitem{Attention2}
Zhao, T., Wu, X.: Pyramid feature attention network for saliency detection. In:
  CVPR (2019)

\end{thebibliography}
\end{document}